\documentclass[11pt,a4paper,table]{article}
\usepackage[hyperref]{AACL-IJCNLP2020}
\usepackage{times}
\usepackage{latexsym}

\usepackage{graphicx}
\usepackage{enumitem}
\usepackage[explicit]{titlesec}
\usepackage{url}
\usepackage{microtype}

\usepackage[noorphans,vskip=0ex]{quoting}

 \titlespacing{\paragraph}{%
   0em}{
   1ex}{
  0\baselineskip}%
\usepackage{xparse}
\NewDocumentCommand\cryingFace{}{\includegraphics[scale=0.1]{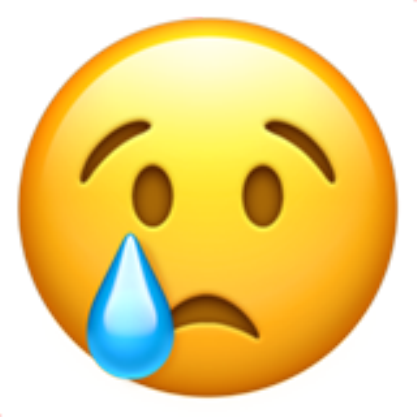}}
\NewDocumentCommand\blackHeart{}{\includegraphics[scale=0.1]{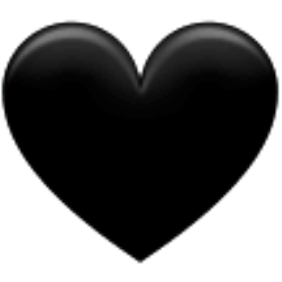}}
\NewDocumentCommand\palmTree{}{\includegraphics[scale=0.1]{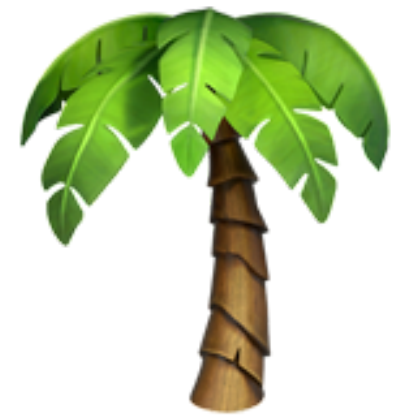}}
\NewDocumentCommand\moyai{}{\includegraphics[scale=0.1]{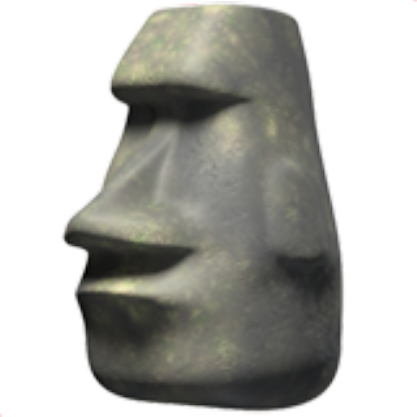}}
\NewDocumentCommand\flower{}{\includegraphics[scale=0.1]{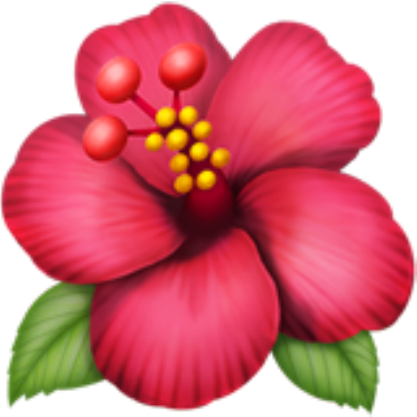}}
\NewDocumentCommand\tulip{}{\includegraphics[scale=0.08]{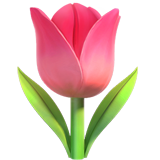}}
\NewDocumentCommand\hamburger{}{\includegraphics[scale=0.08]{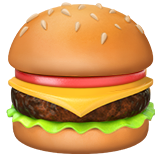}}
\NewDocumentCommand\pizza{}{\includegraphics[scale=0.08]{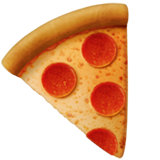}}
\NewDocumentCommand\deli{}{\includegraphics[scale=0.08]{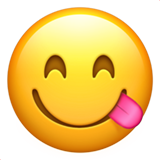}}
\NewDocumentCommand\twobeers{}{\includegraphics[scale=0.08]{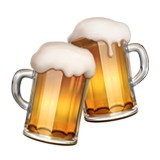}}
\NewDocumentCommand\beer{}{\includegraphics[scale=0.08]{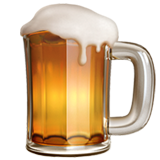}}
\NewDocumentCommand\bottle{}{\includegraphics[scale=0.08]{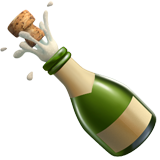}}
\NewDocumentCommand\megaphone{}{\includegraphics[scale=0.08]{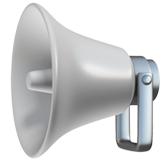}}
\NewDocumentCommand\wave{}{\includegraphics[scale=0.08]{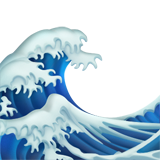}}
\NewDocumentCommand\plane{}{\includegraphics[scale=0.08]{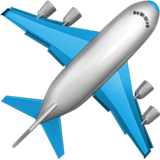}}

\aclfinalcopy 

\title{Point-of-Interest Type Inference from Social Media Text}

\author{
    {\bf Danae S\'{a}nchez Villegas$^\alpha$} \quad {\bf Daniel Preo\c{t}iuc-Pietro$^\beta$} \quad {\bf Nikolaos Aletras$^\alpha$}\\
    $^\alpha$ Computer Science Department, University of Sheffield, UK\\
    $^\beta$ Bloomberg\\
    {\small
    {\tt \{dsanchezvillegas1, n.aletras\}@sheffield.ac.uk}}\\
    {\small
    {\tt dpreotiucpie@bloomberg.net}}
}

\date{}

\begin{document}
\maketitle
\begin{abstract}
Physical places help shape how we perceive the experiences we have there. We study the relationship between social media text and the type of the place from where it was posted, whether a park, restaurant, or someplace else. To facilitate this, we introduce a novel data set of $\sim$200,000 English tweets published from 2,761 different points-of-interest in the U.S., enriched with place type information. We train classifiers to predict the type of the location a tweet was sent from that reach a macro F1 of 43.67 across eight classes and uncover the linguistic markers associated with each type of place. The ability to predict semantic place information from a tweet has applications in recommendation systems, personalization services and cultural geography.\footnote{Data is available here: \url{https://archive.org/details/poi-data}}

\end{abstract}

\section{Introduction}



Social networks such as Twitter allow users to share information about different aspects of their lives including feelings and experiences from places that they visit, from local restaurants to sport stadiums and parks. Feelings and emotions triggered by performing an activity or living an experience in a Point-of-Interest (POI) can give a glimpse of the atmosphere in that place~\citep{Tanasescu2013}.

In particular, the language used in posts from POIs is an important component that contributes toward the place's identity and has been extensively studied in the context of social and cultural geography~\citep{Tuan1991,Scollon2003,Benwell2006}. Social media posts from a particular location are usually focused on the person posting the content, rather than on providing explicit information about the place. Table~\ref{tab:examples} displays example Twitter posts from different POIs. Users express their feelings related to a certain place (`this places gives me war flashbacks'), comments and thoughts associated with the place they are in (`few of us dressed appropriately') or activities they are performing (`leaving the news station', `on the way to the APCE Annual').


In this paper, we aim to study the language that people on Twitter use to share information about a specific place they are visiting.
Thus, we define the prediction of a POI type given a post (i.e. tweet) as a multi-class classification task using only information available at posting time. Given the text from a user's post, our goal is to predict the correct type of the location it was posted, e.g. park, bar or shop. Inferring the type of place from a user's post using linguistic information, is useful for cultural geographers to study a place's identity~\citep{Tuan1991} and has downstream geosocial applications such as POI visualisation~\citep{Mckenzie2015} and recommendation~\citep{Alazzawi2012,Yuan2013,preotiuc2013mining,Gao2015}.


Predicting the type of a POI is inherently different to predicting the POI type from comments or reviews. The role of the latter is to provide opinions or descriptions of the places, rather than the activities and feelings of the user posting the text~\citep{Mckenzie2015}, as illustrated in Table~\ref{tab:examples}. This is also different, albeit related, to the popular task of geolocation prediction~\cite{cheng2010,eisenstein-etal-2010-latent,han-etal-2012-geolocation,roller-etal-2012-supervised,rahimi-etal-2015-exploiting,dredze-etal-2016-geolocation}, as this aims to infer the exact geographical location of a post using language variation and geographical cues rather than inferring the place's type. Our task aims to uncover the geographic agnostic features associated with POIs of different types.

\renewcommand{\arraystretch}{1.2}
\begin{table*}[t!]

\centering
\resizebox{\linewidth}{!}{
\begin{tabular}{ l l r r r r}
\hline
\rowcolor[HTML]{C0C0C0} 
\textbf{Category}  & \textbf{Sample Tweet} & \textbf{Train} & \textbf{Dev} & \textbf{Test} & \textbf{Tokens} \\ \hline
\textbf{Arts \& Entertainment}        & i'm back in central park . this place gives me war flashbacks now lol                                                                                   & 40,417 & 4,755 & 5,284 & 14.41      \\ \hline
\rowcolor[HTML]{EFEFEF} 
\textbf{College \& University}        & currently visiting my dream school \cryingFace{}{} \blackHeart{}{}                                                                                                       & 21,275 & 2,418 & 2,884  & 15.52     \\ \hline
\textbf{Food}                         & Some Breakfast, it's only right! \#LA                                                                                                        & 6,676 & 869 & 724   & 14.34      \\ \hline
\rowcolor[HTML]{EFEFEF} 
\textbf{Great Outdoors}               & \begin{tabular}[c]{@{}l@{}}Sorry Southport, Billy is dishing out donuts at \#donutfest today. See you\\ next weekend!\end{tabular}  & 27,763 & 4,173 & 3,653 & 13.49 \\ \hline
\textbf{Nightlife Spot}               & \begin{tabular}[c]{@{}l@{}}Chicago really needs to step up their Aloha shirt game. Only a few of us\\ dressed ``appropriately" tonight. :) \moyai{}{} \palmTree{ \flower{}{}}{}\end
{tabular} & 5,545 & 876 & 656 & 15.46 \\ \hline
\rowcolor[HTML]{EFEFEF} 
\textbf{Professional \& Other Places} & Leaving the news station after a long day                                                                               & 30,640 & 3,381 & 3,762 & 16.46 \\ \hline
\textbf{Shop \& Service}              & Came to get an old fashioned tape measures and a button for my coat                                                     & 8,285 & 886 & 812 & 15.31    \\ \hline
\rowcolor[HTML]{EFEFEF} 
\textbf{Travel \& Transport}          & \begin{tabular}[c]{@{}l@{}}Shoutout to anyone currently on the way to the APCE Annual Event in\\ Louisville, KY! \#APCE2018\end{tabular}     & 16,428 & 2,201 & 1,872 & 14.88       \\ \hline
\end{tabular}
}
\caption{Place categories with sample tweets and data set statistics.}
\label{tab:examples}
\end{table*}


Our contributions are as follows: (1) We provide the first study of POI type prediction in computational linguistics; (2) A large data set made out of tweets linked to particular POI categories; (3) Linguistic and temporal analyses related to the place the text was posted from; (4) Predictive models using text and temporal information reaching up to 43.67 F1 across eight different POI types.

\section{Point-of-Interest Type Data}
We define the POI type prediction as a multi-class classification task performed at the social media post level. Given a post T, defined as a sequence of tokens $T =\{t_1,...,t_n\}$, the goal is to label T as one of the $M$ POI categories. We create a novel data set for POI type prediction containing text and the location type it was posted from as, to the best of our knowledge, no such data set is available. We use Twitter as our data source because it contains a large variety of linguistic information such as expression of thoughts, opinions and emotions~\cite{java2007we,kouloumpis2011twitter}.

\subsection{Types of POIs}
Foursquare is a location data platform that manages `Places by Foursquare', a database of more than 105 million POIs worldwide. The place information includes verified metadata such as name, geo-coordinates and categories as well as other user-sourced metadata such as tags, comments or photos. POIs are organized into 9 top level primary categories with multiple subcategories. We only focus on 8 primary top-level POI categories since the category `Residence' has a considerably smaller number of tweets compared to the other categories (0.78\% tweets from the total). We leave finer-grained place category inference as well as using other metadata for future work since the scope of this work is to study the language of posts associated with semantic type places.

\subsection{Associating Tweets with POI Types}
Twitter users can tag their tweets to the locations they are posted from by linking to Foursquare places.\footnote{\url{https://developer.foursquare.com/places}} 
In this way, we collect tweets assigned to the POIs and associated metadata (see Table~\ref{tab:examples}). We select a broad range of locations for our experiments. There is no public list of all Foursquare locations that can be used through Twitter and can be programmatically accessed. Hence, in order to discover Foursquare places that are actually used in tweets, we start with all places found in a 1\% sample of the Twitter feed between 31 July 2016 and 24 January 2017 leading us to a total of 9,125 different places. Then, we collect all tweets from these places between 17 August 2016 and 1 March 2018 using the Twitter Search API\footnote{\url{https://developer.twitter.com/en/docs/tweets/search/guides/tweets-by-place}}. We collect the place metadata from the public Foursquare Venues API. This resulted in a total data set of 1,648,963 tweets tagged to a Foursquare place. In order to extract metadata about each location, we crawled the Twitter website to identify the corresponding Foursquare Place ID of each Twitter place. Then, we used the public Foursquare Venues API\footnote{\url{https://developer.foursquare.com/overview/venues.html}} to download all the place metadata. 

\subsection{Data Filtering}
To limit variation in our data, we filter out all non-English tweets and non-US places, as these were very limited in number. We keep POIs with at least 20 tweets and randomly subsample 100 tweets from POIs with more tweets to avoid skewing our data. Our final data set consists of 196,235 tweets from 2,761 POIs. 








\subsection{Data Split}
We create our data split at a location-level to ensure that our models are robust and generalize to locations held-out in training. We split the locations in train (80\%), development (10\%) and test (10\%) sets and assign tweets to one of the three splits based on the location they were posted from (see Table \ref{tab:examples} for detailed statistics).

\subsection{Text Processing}
We lower-case text and replace all URLs and mentions of users with placeholders. We preserve emoticons and punctuation and replace tokens that appear in less than five tweets with an `unknown' token. We tokenize text using a Twitter-aware tokenizer~\cite{schwartz-etal-2017-dlatk}.

\begin{table*}[t!]
\centering
\resizebox{\linewidth}{!}{
\begin{tabular}{|l|l|l|l|l|l|l|l|l|l|l|l|l|l|l|l|}
\hline
\rowcolor[HTML]{9B9B9B} 
\multicolumn{2}{|c|}{\textbf{Arts}} & \multicolumn{2}{|c|}{\textbf{College}} & \multicolumn{2}{|c|}{\textbf{Food}} & \multicolumn{2}{|c|}{\textbf{Outdoors}} & \multicolumn{2}{|c|}{\textbf{Nightlife}} & \multicolumn{2}{|c|}{\textbf{Professional}} & \multicolumn{2}{|c|}{\textbf{Shop}} & \multicolumn{2}{|c|}{\textbf{Travel}}  \\ \hline
\rowcolor[HTML]{C0C0C0} 
Feature & \textbf{$\chi^2$} & Feature & \textbf{$\chi^2$} & Feature & \textbf{$\chi^2$} & Feature & \textbf{$\chi^2$} & Feature & \textbf{$\chi^2$} & Feature & \textbf{$\chi^2$} & Feature & \textbf{$\chi^2$} & Feature & \textbf{$\chi^2$}
\\ \hline
concert & 167.20 & campus & 298.74 & chicken & 375.52 & beach & 591.81 & \#craftbeer & 425.97 & school & 87.46 & mall & 462.03 & airport & 394.20  \\ \hline
\rowcolor[HTML]{EFEFEF} 
museum & 152.14 & college & 266.63 & \#nola & 340.64 & \wave{}{} & 239.00 & \twobeers{}{} & 311.68 & students & 79.93 & store & 403.00 & \plane{}{} & 343.30  \\ \hline
show & 134.39 & university & 155.65 & lunch & 255.98 & hike & 227.91 & beer & 203.57 & grade & 66.05 & shopping & 359.00 & flight & 292.94  \\ \hline
\rowcolor[HTML]{EFEFEF} 
night & 104.48 & class & 112.23 & fried & 216.49 & lake & 193.58 & bar & 93.90 & vote & 65.80 & shop & 132.39 & hotel & 168.38  \\ \hline
tonight & 80.76 & semester & 103.19 & dinner & 203.65 & park & 165.92 & \beer{}{} & 67.00 & our & 63.12 & \tulip{}{} & 126.07 & conference & 141.74  \\ \hline
\rowcolor[HTML]{EFEFEF} 
game & 73.56 & football & 59.24 & \hamburger{}{} & 195.41 & island & 151.45 & \bottle{}{} & 56.94 & jv & 60.64 & \megaphone{}{} & 95.32 & landed & 118.05  \\ \hline
art & 69.77 & student & 57.86 & pizza & 190.83 & sunset & 142.44 & dj & 56.56 & church & 52.97 & apple & 88.74 & plane & 88.42  \\ \hline
\rowcolor[HTML]{EFEFEF} 
USER & 66.14 & classes & 57.37 & shrimp & 188.77 & hiking & 137.74 & tonight & 53.39 & hs & 50.63 & market & 76.60 & bound & 78.43  \\ \hline
zoo & 66.09 & students & 56.98 & \pizza{}{} & 179.39 & beautiful & 109.45 & ale & 52.62 & senior & 50.05 & auto & 73.52 & heading & 62.09  \\ \hline
\rowcolor[HTML]{EFEFEF} 
baseball & 62.90 & camp & 44.19 & \deli{}{} & 151.00 & bridge & 108.56 & party & 51.14 & ss & 44.46 & stock & 72.31 & headed & 57.12  \\ \hline
\end{tabular}
}
\caption{Unigrams associated with each category, sorted by $\chi^2$ value computed between the normalized frequency of each feature and the category label across all tweets in the training set ($p<0.001$).}
\label{tab:linguisticanalysis}
\end{table*}

\section{Analysis}
We first analyze our data set to understand the relationship between location type, language and posting time.


\subsection{Linguistic Analysis}
We analyze the linguistic features specific to each category by ranking unigrams that appear in at least 5 different locations, such that these are representative of the larger POI category rather than a few specific places. Features are normalized to sum up to unit for each tweet, then we compute the (Pearson) $\chi^2$ coefficient independently between its distribution across posts and the binary category label of the post similar to the approach followed by \citet{Maronikolakis2020} and \citet{Preotiuc2019complaints}. Table~\ref{tab:linguisticanalysis} presents the top unigram features for each category.

We note that most top unigrams specific of a category naturally refer to types of places (e.g. `campus', `beach', `mall', `airport') that are part of that category. All categories also contain words that refer to activities that the poster of the tweet is performing or observing while at a location (e.g. `camp' and `football' for College, `concert' and `show' for Arts \& Entertainment, `party' for Nightlife Spot, `landed' for Travel \& Transport, `hike' for Greater Outdoors). Nightlife Spot and Food categories are represented by types of food or drinks that are typically consumed at these locations. Beyond these typical associations, we highlight that usernames are more likely mentioned in the Arts \& Entertainment category, usually indicating activities involving groups of users, emojis indicative of the user state (e.g. happy emoji in Food places) and adjectives indicative of the user's surroundings (e.g. `beautiful' in Greater Outdoors places). Finally, we also uncover words indicative of the time the user is at a place, such as `tonight' for Arts \& Entertainment, `sunset' for the Greater Outdoors and `night' for Nightlife Spots and Arts \& Entertainment.

\subsection{Temporal Analysis}
\begin{figure}[t!]
     \centering
     \includegraphics[width=0.9\columnwidth]{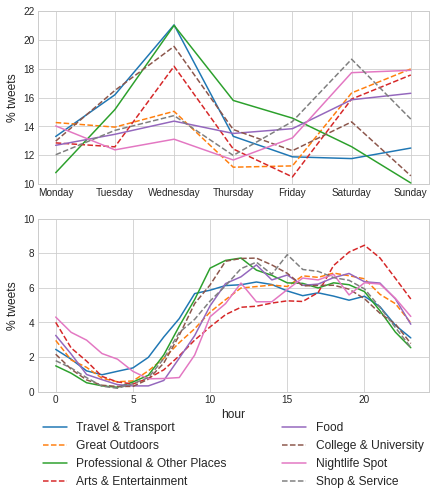}
     \caption{Percentage of tweets by day of week (top) and by hour of day (bottom).}
\label{fig:dow}
\end{figure}

We further examine the relationship between the time a tweet was posted and the POI type it was posted from. Figure~\ref{fig:dow} shows the percentage of tweets by day of week (top) and hour of day (bottom). 

We observe that tweets posted from the `Professional \& Other Places', `Travel \& Transport' and `College \& University' categories are more prevalent on weekdays, peaking on Wednesday, while on weekends more tweets are posted from the `Great Outdoors', `Arts \& Entertainment', `Nightlife \& Spot' and `Food' categories when people focus less on professional activities and dedicate more time to leisure as expected. The hour of day pattern follows the daily human activity rhythm, but the differences between categories are less prominent, perhaps with the exception of the `Arts \& Entertainment' category peaks around 8PM and `Nightlife Spots' that see a higher percent of tweets in the early hours of the day (between 1-5am) than other categories.

\section{Predicting POI Types of Tweets}


\subsection{Methods}

\paragraph{Logistic Regression}
~We first experiment with logistic regression using a standard bag of n-grams representation of the tweet (\textbf{LR-W}), including unigrams to trigrams weighted using TF-IDF. We identified in the analysis section that temporal information about the tweet may be useful for classification. Hence, to add temporal information extracted from a tweet, we create a 31-dimensional vector encoding the hour of the day and the day of the week it was sent from. We experiment with only using the temporal features (\textbf{LR-T}) and in combination with the text features (\textbf{LR-W+T}). We use L1 regularization \cite{Hoerl1970} with hyperparameter $\alpha = .01$ (selected based on dev set from \{.001, .01, .1\}).

\paragraph{BiLSTM}
~We train models based on bidirectional Long-Short Term Memory (LSTM) networks \cite{hochreiter1997long}, which are popular in text classification tasks. Tokens in a tweet are mapped to embeddings and passed through the two LSTM networks, each processing the input in opposite directions. The outputs are concatenated and passed to the output layer using a softmax activation function (\textbf{BiLSTM}). We extend the BiLSTM to encode temporal one-hot representation by: (a) concatenating the temporal vector to the tweet representation (\textbf{BiLSTM-TC}); and (b) projecting the time vector into a dense representation using a fully connected layer which is added to the tweet representation before passing it through the output layer using a softmax activation function (\textbf{BiLSTM-TS}). We use 200-dimensional GloVe embeddings \cite{pennington-etal-2014-glove} pre-trained on Twitter data. The maximum sequence length is set to 26, covering 95\% of the tweets in the training set. The LSTM size is $h$ = 32 where $h \in \{32,64,100,300\}$ with dropout $d$ = 0.5 where $d \in \{.2,.5\}$. We use Adam \cite{kingma2014adam} with default learning rate, minimizing cross-entropy using a batch size of 32 over 10 epochs with early stopping. 


\paragraph{BERT}
~Bidirectional Encoder Representations from Transformers (BERT) is a pre-trained language model based on transformer networks~\cite{Vaswani2017, devlin-etal-2019-bert}. BERT consists of multiple multi-head attention layers to learn bidirectional embeddings for input tokens. The model is trained on masked language modeling, where a fraction of the input tokens in a given sequence is replaced with a mask token, and the model attempts to predict the masked tokens based on the context provided by the non-masked tokens in the sequence. We fine-tune BERT for predicting the POI type of a tweet by adding a classification layer with softmax activation function on top of the Transformer output for the `classification' $[CLS]$ token (\textbf{BERT}). Similarly to the previous models, we extend BERT to make use of the time vector in two ways, by concatenating (\textbf{BERT-TC}), and by adding it (\textbf{BERT-TS}) to the output of the Transformer before passing it to through the classification layer with softmax activation function. We use the base model (12-layer, 110M parameters) trained on lower-cased English text. We fine-tune it for 2 epochs with a learning rate $l=2e^{-5}$, $l \in \{2e^{-5}, 3e^{-5}, 5e^{-5}\}$ and a batch size of 32.

\subsection{Results}

Table~\ref{tab:results} presents the results of POI type prediction measured using accuracy, macro F1, precision and recall across three runs. In general, we observe that we can predict POI types of tweets with good accuracy, considering the classification is across eight relatively well balanced classes.

Best results are obtained using BERT-based models (BERT, BERT-TC and BERT-TS), with the highest accuracy of 49.17 (compared to 26.89 majority class) and highest macro-F1 of 43.67 (compared to 12.64 random). We observe that BERT models outperform both BiLSTM and linear methods across all metrics, with over 4\% improvement in accuracy and 5 points F1. The BiLSTM models perform marginally better than the linear models. Temporal features alone are marginally useful when models are evaluated using accuracy (+0.28 BERT, +0.34 for BiLSTMs, +0.69 for LR) and perform similarly on F1, with the notable exception of the BiLSTM models. We find that adding these features is more beneficial than concatenating them, with concatenation hurting performance on accuracy for both BiLSTM and BERT. 

\begin{figure}[t!]
     \centering
     \includegraphics[width=\columnwidth]{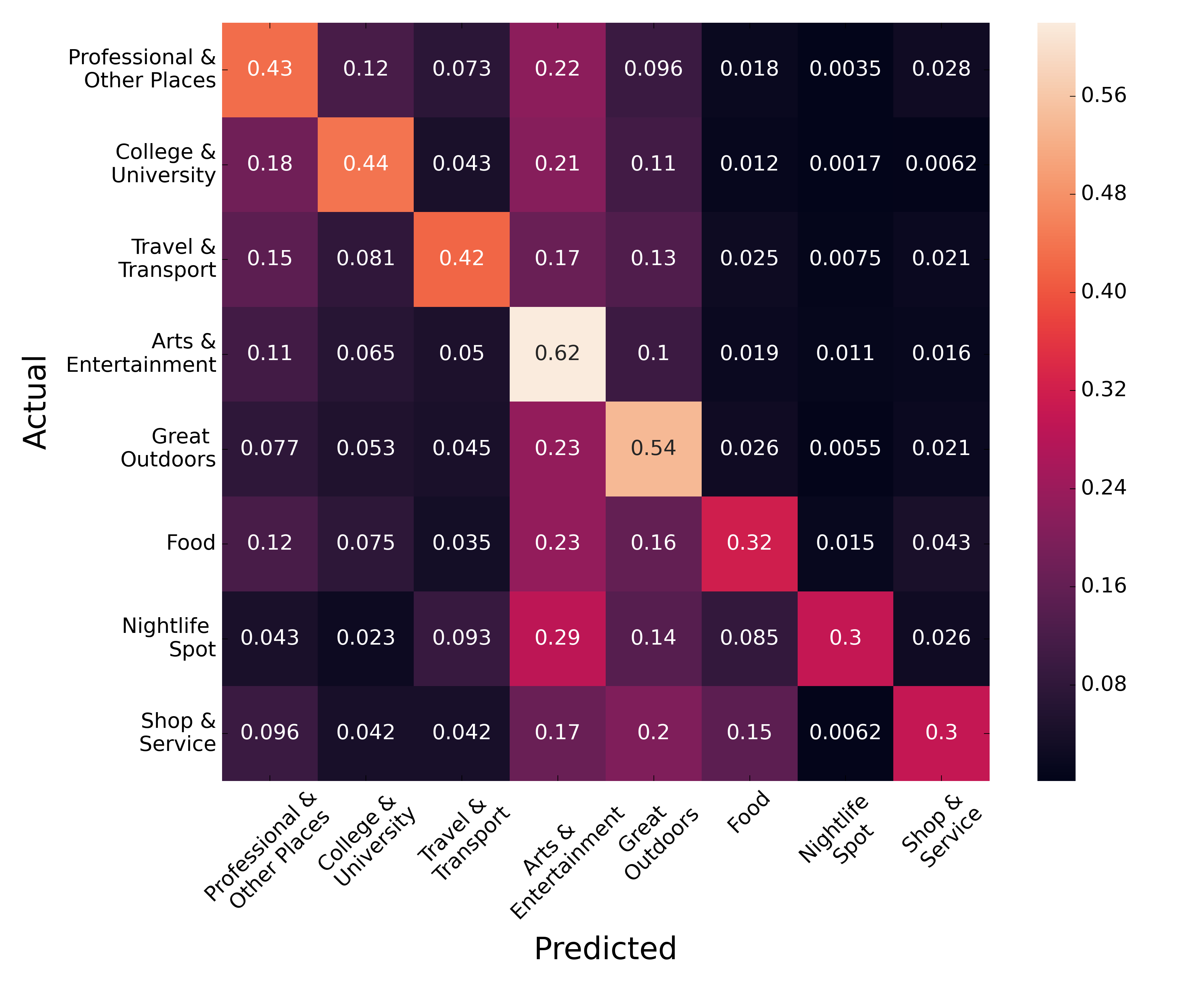}
     \caption{Confusion Matrix of the best performing model (BERT).}
\label{fig:confusion_bert}
\end{figure}

Figure~\ref{fig:confusion_bert} shows the confusion matrix of our best performing model, BERT, according to the macro-F1 score. The confusion matrix is normalized over the actual values (rows). The category `Arts \& Entertainment` has the greatest percentage (62\%) of correctly classified tweets, followed by the `Great Outdoors` category with 54\%, and the  `College \& University` category with 44\%. On the other hand, the categories `Nightlife Spot` and `Shop \& Service` have the lowest results, where 30\% of the tweets predicted as each of these classes is correctly classified. Most common error is when the model classifies tweets from the category `College \& University' as `Professional \& Other Places', as tweets from these places contain similar terms such as `students' or `class'.

\renewcommand{\arraystretch}{1.2}
\begin{table}[]
\centering
\footnotesize
\begin{tabular}{|
>{\columncolor[HTML]{EFEFEF}}l |c|c|c|c|}
\hline
\multicolumn{1}{|c|}{\cellcolor[HTML]{9B9B9B}\textbf{Model}} & \cellcolor[HTML]{9B9B9B}\textbf{Acc} & \cellcolor[HTML]{9B9B9B}\textbf{F1} & \cellcolor[HTML]{9B9B9B}\textbf{P} & \cellcolor[HTML]{9B9B9B}\textbf{R} \\ \hline
\textbf{Major. Class}                                        & 26.89                                & 5.30                                & 3.36                               & 12.50                              \\ 
\textbf{Random}                                              & 13.63                                & 12.64                               & 13.63                              & 15.68                              \\ \hline
\textbf{LR-T}                                                & 27.93                                & 14.01                               & 15.78                              & 16.06                              \\ 
\textbf{LR-W}                                              & 43.04                                & 37.33                               & 37.06                              & 38.03                              \\ 
\textbf{LR-W+T}                                               & 43.73                                & 37.83                               & 37.68                              & 38.37                              \\ \hline
\textbf{BiLSTM}                                              & 44.38                                & 35.77                               & 45.29                              & 33.78                              \\ 
\textbf{BiLSTM-TC}                                           & 44.01                                & 38.07                               & 41.51                              & 36.46                              \\ 
\textbf{BiLSTM-TS}                                           & 44.72                                & 38.26                               & 42.91                              & 36.30                              \\ \hline
\textbf{BERT}                                                & 48.89                                & \textbf{43.67}                               & \textbf{48.44}                              & \textbf{41.33 }                             \\ 
\textbf{BERT-TC}                                             & 46.13                                & 41.19                               & 46.81                              & 39.03                              \\ 
\textbf{BERT-TS}                                             &\textbf{ 49.17 }                               & 43.47                               & 48.40                              & 41.26                              \\ \hline
\end{tabular}
\caption{Accuracy (Acc), Macro-F1 Score (F1), Precision macro (P), and Recall macro (R) for POI type prediction (all std. dev $<$ 0.01). Best results are in bold.}
\label{tab:results}
\end{table}

\section{Conclusion}

We presented the first study on predicting the POI type a social media message was posted from and developed a large-scale data set with tweets mapped to their POI category. We conducted an analysis to uncover features specific to place type and trained predictive models to infer the POI category using only tweet text and posting time with accuracy close to 50\% across eight categories. Future work will focus on using other modalities such as network~\cite{Aletras2018,Tsakalidis2018} or image information~\cite{vempala-preotiuc-pietro-2019-categorizing,alikhani-etal-2019-cite} and prediction at a more granular level of POI types.

\section*{Acknowledgments}

DSV is supported by the Centre for Doctoral Training in Speech and Language Technologies (SLT) and their Applications funded by the UK Research and Innovation grant EP/S023062/1. NA is supported by ESRC grant ES/T012714/1.



\bibliographystyle{acl_natbib}
\bibliography{poi2020,anthology}

\end{document}